\documentclass[twocolumn]{svjour3}

\usepackage[utf8]{inputenc}
\usepackage[accsupp]{axessibility}
\usepackage{listings}
\usepackage[ruled,vlined]{algorithm2e}
\usepackage{graphicx}
\usepackage{booktabs}
\usepackage{multirow}
\usepackage[export]{adjustbox}
\usepackage{tikz}
\usepackage{amsmath}
\usepackage{amssymb}
\usepackage{floatflt}
\usepackage[table]{xcolor}
\usepackage{color}
\usepackage{wrapfig}
\usepackage{times}
\usepackage{epsfig}
\usepackage{epstopdf}
\usepackage{array}
\usepackage{pifont}
\usepackage{sidecap}
\usepackage{tablefootnote}
\usepackage{threeparttable}
\usepackage[normalem]{ulem}
\usepackage{enumitem}
\usetikzlibrary{calc}

\newcommand{\A}{\mathbf{A}}

\useunder{\uline}{\ul}{}
\newcolumntype{L}[1]{>{\raggedright\arraybackslash}p{#1}}
\newcolumntype{C}[1]{>{\centering\arraybackslash}p{#1}}
\newcolumntype{R}[1]{>{\raggedleft\arraybackslash}p{#1}}
\usepackage[numbers,sort&compress]{natbib}
\usepackage{titlesec}
\definecolor{cvprblue}{rgb}{0.21,0.49,0.74}
\usepackage[breaklinks,colorlinks,allcolors=cvprblue]{hyperref}
\usepackage{orcidlink}
\usepackage{array}
\usepackage{amssymb}
\DeclareUnicodeCharacter{200B}{}
\usepackage{pifont}
\newcommand{\cmark}{\ding{51}}
\newcommand{\xmark}{\ding{55}}

\begin{document}

\title{Self-Improving is Often Sudden:  Enlightenment-style Finetuning for Large-Scale Models}

\titlerunning{Enlightenment-style Tuning for Large Models}

\author{Jing-Xiao Liao\textsuperscript{1} \and
Tianwei Zhang\textsuperscript{2} \and
Yu-Hao Jiang\textsuperscript{1} \and
Feifei Zhang\textsuperscript{3}
\and
Hang-Cheng Dong\textsuperscript{4} \and
Feng-Lei Fan\textsuperscript{1,$\star$}}

\authorrunning{JX Liao et al.}

\institute{\textsuperscript{1}Frontier of Artificial Network Lab, Department of Data Science, City University of Hong Kong, Hong Kong SAR, China \and
\textsuperscript{2}The Shenzhen Institute of Artificial Intelligence and Robotics for Society, Shenzhen, China, The Chinese University of Hong Kong-Shenzhen, Shenzhen, China
\and
\textsuperscript{3}School of Computer Science and Artificial Intelligence,
Guangdong University of Education, Guangzhou 510303, China \and
\textsuperscript{4}School of Instrumentation Science and Engineering, Harbin Institute of Technology, Harbin, China. \\
\textsuperscript{$\star$}Corresponding author.}

\date{Received: date / Accepted: date}

\maketitle

\begin{abstract}
The pursuit of autonomously self-improving models has attracted growing interest in the era of large-scale foundation models. Drawing inspiration from the concept of ``enlightenment'' or “aha moment” in human brain, we hypothesize that large models exhibit an analogous enlightenment phenomenon — a latent capacity for sudden capability boost. Then, we propose \texttt{Enlightenment}, a novel training-free post-tuning paradigm for large-scale models. Our approach modifies shortcuts for key modules/layers without weight updates, while existing training-free ones predominantly manipulate attention weights. We introduce two architecture-specific instantiations: i) For large language models, we propose attention head-mixing shortcuts that recalibrate attention weights by linking the initial attention head's output to all other target heads, modulated by an adaptive scaling factor initialization strategy. ii) For vision-language models, we apply a lightweight scalar-modulated factor to residual connections in the decoder layers, regulating information flow. Extensive experiments show that \texttt{Enlighten}
\texttt{-ment} efficiently unlocks the latent potential of pre-trained networks, yielding remarkable performance improvements across diverse benchmarks and models. {We have open-sourced our code at \href{https://github.com/asdvfghg/Enlightenment_project}{Enlightenment} for readers' free check.} 
\end{abstract}

\section{Introduction}
\label{sec1}
\vspace{-0.2cm}

Recently, enabling a large-scale model to improve itself has gained large traction in the AI community~\cite{wu2025meta,qu2024recursive,huang2023large}. This venture actually can be found across the history of AI—from early symbolic self‑modification to the self‑play dynamics that powered breakthroughs in games like chess, and beyond~\cite{billard2002symbolic,udrescu2020ai}. With the rapid advancement of large-scale pretrained models, realizing their effective self-improvement has emerged as a pivotal challenge on the path toward artificial general intelligence (AGI), promising to endow these models with an unprecedented degree of adaptability and generalizability. However, for this moment, self-improvement is a research area whose definition and boundary are not crystal clear. 

\begin{figure*}[ht]
    \centering
\includegraphics{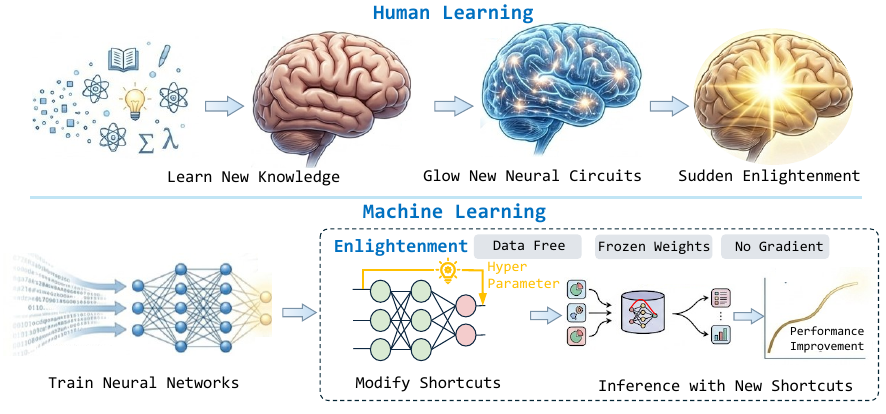}
    \caption{The overall concept of Enlightenment. Just as sudden comprehension arises from the formation of new neuronal connections in the human brain, we modify shortcuts within pre-trained models, enabling them to self-improve during inference without additional post-training.}
    \label{fig:overall}
    \vspace{-0.5cm}
\end{figure*}

To the best of our knowledge, self-improvement in contemporary research mainly concerns four schools: i) Continual learning provides the machinery for models to adapt to new tasks without catastrophic forgetting, employing elastic weight consolidation, dynamic architecture expansion, or episodic memory replay so that each time of learning builds upon previous knowledge rather than erasing it~\cite{ning2026sparse}; ii) Self-distillation leverages the model as its own oracle: the model is iteratively retrained on softened distributions produced by its own earlier checkpoints, which progressively sharpens predictions without external ground truth~\cite{gou2023multi,jiang2026teacher}; iii) Test-time adaptation, in contrast, updates model statistics or a small subset of parameters on a per-sample or per-distribution basis using unsupervised objectives like entropy minimization, pseudo-labeling, or self-supervised rotation prediction, which allows the model to rapidly conforming to individual user patterns after deployment~\cite{liang2025comprehensive}; iv) Training-free finetuning explores token-level attention modulation for improving foundation model performance without any parameter modifications~\cite{han2026zerotuning, xiao2024efficient}.

It can be seen that a diverse spectrum of approaches are related to model self-improvement. Yet, no single methodology has yet crystallized into a dominant trajectory. Our intention here is not to reconcile or unify these existing routes. Rather, given the multi-faceted nature of self-improvement, we take a step back and design a new mechanism in a principled way, thereby offering a perspective that is inherently distinct from existing lines. We turn our attention to the roots of brain intelligence and seek inspiration from our brain. This aligns with the idea of “NeuroAI" \cite{zador2022toward,sadeh2025emergence, fan2025towards} which argues that insights from neuroscience can greatly accelerate the development of next‑generation AI. This claim is well grounded: the brain remains the most intelligent system known, and artificial neural networks can be viewed as simplified abstractions of it. Thus, mechanisms of neuroscience can more or less shed light on the challenges in AI. 

Motivated by the NeuroAI paradigm, it is noted that human's self-improving is often sudden, manifesting as the so-called “aha moment" or enlightenment. In the human brain, enlightenment is not merely subjective flashes of clarity. Neuroimaging and electrophysiological studies suggest that such insights often involve the abrupt recruitment of the anterior superior temporal gyrus and the anterior cingulate cortex, accompanied by a transient burst of gamma‑band activity~\cite{arikha2006form}. In addition, enlightenment often appears during resting time when active learning or training has ceased. This resting-time‑driven plasticity change can sometimes establish entirely new microcircuits, thereby enabling a qualitative leap in performance rather than incremental adjustment~\cite{monsivais2017seasonal}. Given the compelling characteristics of enlightenment, \textit{can we introduce the related mechanism into large models and examine the associated computational benefits in self-improvement?}

Translated to artificial systems, enlightenment suggests that a new operational phase in manipulating models beyond training and testing, referred to as retrospection, is important, which tunes models in the resting time and not necessarily relying on data. More importantly, the model may wire new circuits to improve itself during retrospection~\cite{chen2017computational}. Furthermore, we posit that enlightenment in artificial networks should fulfill the following conditions: i) no gradient is computed, since we highlight the sudden change; ii) no data is involved, since we highlight the resting-time adjustment that is before testing and after training; and iii) no weight update is needed, since we highlight modifying circuits. Per these conditions, we propose \texttt{Enlightenment}, a novel and highly effective training-free retrospection method for large models. As Figure \ref{fig:overall} shows, we focus on modifying the network structure, with an emphasis on shortcuts, and we even inject connections that do not exist previously into large-scale models. Table \ref{tab:comparsion} shows that the proposed \texttt{Enlightenment} occupies a unique niche relative to other techniques. Particularly, the existing training-free finetuning methods focus on plasticity adjustments, while our method emphasizes internal structures. 

\begin{table*}[ht]
\label{tab:comparsion}
\caption{Comparisons of different self-improvement paradigms.}
\begin{tabular}{@{}lccccc@{}}
\toprule
\textbf{Dimension}        & \textbf{Continual Learning}                                                                                         & \textbf{Self-Distillation}                                                                                                    & \textbf{Test-Time Adaptation}                                                                                                       & \textbf{Training-Free Finetuning}                                                                                 & \textbf{Enlightenment (Ours)} \\ \midrule
\textbf{New Tasks}        &       {\color[HTML]{32CB00}\cmark}                                                                                                              &           \xmark                                                                                                                    &                                                       \xmark                                                                              &                 \xmark                                                                                                  &         \xmark                       \\
\textbf{New Weights}      &        \xmark                                                                                                         &          {\color[HTML]{32CB00}\cmark}                                                                                                                    &     {\color[HTML]{32CB00}\cmark}                                                                                                                                  &           {\color[HTML]{32CB00}\cmark}                                                                                                          &      \xmark                          \\
\textbf{New Structures}   &      \xmark                                                                                                               &  {\color[HTML]{32CB00}\cmark}                                                                                                                               & \xmark                                                                                                                                    &                \xmark                                                                                                   &              {\color[HTML]{32CB00}\cmark} \\

\textbf{New Phase}   &      \xmark                                                                                 &  \xmark                                                                                                                       & \xmark                                                                                                                                    &                  {\color[HTML]{32CB00}\cmark}                                                                                                  &              {\color[HTML]{32CB00}\cmark} \\

\midrule
\textbf{Data Free}     & {\color[HTML]{FE0000}\xmark} & {\color[HTML]{F8A102}\cmark(partial)} & {\color[HTML]{F8A102}\cmark(partial)}   & {\color[HTML]{32CB00}\cmark}  &     {\color[HTML]{32CB00}\cmark}                 \\
\textbf{Training Free} & {\color[HTML]{FE0000}\xmark}  & {\color[HTML]{FE0000}\xmark}  &   {\color[HTML]{FE0000}\xmark}           &   {\color[HTML]{32CB00}\cmark}  &         {\color[HTML]{32CB00}\cmark}                    \\ \bottomrule
\end{tabular}
\vspace{-0.5cm}
\end{table*}

Specifically, guided by the principle of modifying shortcuts, we implement \texttt{Enlightenment} for VLMs and LLMs, respectively. For VLMs, we apply a single shared scalar modulator to every residual connection within the Transformer decoder layers, covering both the attention and feed-forward network (FFN) sub-blocks. For LLMs, we take a bolder step: inserting new shortcuts to aggregate information across per-head attention outputs from the head to the rest. The efficacy of such intra-layer shortcuts is corroborated by prior work~\cite{11095854,zhang2024intrinsic}. Furthermore, to complement these architectural augmentations, shortcuts are assigned with dynamic scaling hyperparameters, providing a per-head level recalibration. \texttt{Enlightenment} expands the representational capacity of large models without altering pre-existing weights. Meanwhile, our method introduces only a small number of hyperparameters, and the effect of self-improvement is insensitive to most of them. In brief, we summarize our contributions as threefold:

\begin{itemize}
\item[1)] Inspired by the neuroscience of enlightenment, we introduce \texttt{Enlightenment}, a training‑free tuning paradigm that operates in a novel retrospection phase between training and inference. This paradigm focuses on shortcuts instead of weights.

\item[2)] To instantiate this paradigm, we propose a lightweight architectural intervention strategy that modulates shortcut connections via scaling hyperparameters and even inserting cross‑head shortcuts, thereby expanding representational capacity without post‑training.

\item[3)] Extensive experiments on diverse multimodal and reasoning benchmarks demonstrate that \texttt{Enlightenment} consistently delivers salient performance improvements across three widely-used VLM and three mainstream LLM variants, while requiring no additional data, no gradient computation, and no weight updates.
\end{itemize}

\vspace{-0.5cm}

\section{Related Work}

\subsection{Continual Learning}
Continual learning (CL), also referred to as lifelong learning, addresses the challenge of enabling LLMs to acquire new knowledge without forgetting previously learned tasks~\cite{zhou2024continual,chen2026continual}. Continual learning can be regarded as a form of self-improvement. Recently, continual instruction tuning (CIT), which refines a large model's instruction-following ability~\cite{chen2024coin,he2026continual}, has been intensively investigated.  

i) Task-incremental CIT fine-tunes LLMs on sequential task-specific instructions, with methods such as Progressive Prompts~\cite{razdaibiedinaprogressive} freezing most parameters and learning per-task prompt tokens to reduce computation and forgetting, while the scalable language model~\cite{peng2024scalable} incorporates vector space retrieval for knowledge expansion. ii) Domain-incremental CIT adapts LLMs to evolving domains. Task-adaptive pre-training (TAPT) \cite{gururangan2020don} employs a data selection strategy that retrieves in-domain unlabeled text for fine-tuning, and ConPET~\cite{song2023conpet} combines parameter-efficient tuning with dynamic replay to prevent forgetting. iii) Tool-incremental CIT represents a particularly promising direction, extending LLMs beyond text generation to real-world tool use. ToolLLM~\cite{qin2024toolllm} demonstrates that LLMs can master over 16,000 real-world APIs through CIT, enabling appropriate tool selection and invocation for diverse tasks. ToolkenGPT~\cite{hao2023toolkengpt} takes a complementary approach by augmenting frozen language models with tool embeddings, updating only tool-specific parameters to eliminate catastrophic forgetting while continuously expanding tool capabilities.

\vspace{-0.5cm}

\subsection{Self-Distillation}
Self-Distillation (SD) transfers knowledge from a model or its historical states, enabling self-evolution without an external teacher network~\cite{11443225}. In the instruction-following domain, Self-Instruct~\cite{wang2023self} establishes a closed-loop pipeline for autonomously generating, filtering, and refining high-quality fine-tuning data. For preference alignment, self-play fine-tuning (SPIN)~\cite{chen2024self} casts the process as an adversarial self-play game between pseudo-responses and gold-standard SFT trajectories under a direct preference optimization (DPO) objective, breaking reliance on third-party evaluators. Self-distillation fine-tuning (SDFT)~\cite{shenfeld2026self} further addresses manifold contraction from multi-stage fine-tuning by using an undegraded historical model state as a distillation anchor, recovering broad semantic capabilities without sacrificing task-specific performance.

In reasoning domains, SD has enabled breakthroughs in long-chain logical reasoning and code generation. On-policy self-correction (OPSD)~\cite{zhao2026self} trains a single LLM in dual roles—a blind student and a privileged teacher—applying KL or Jensen-Shannon divergence as a probabilistic penalty over on-policy solution trajectories. Simple self-distillation (SSD)~\cite{zhang2026embarrassingly} takes a simpler route, using the model's own unverified temperature-sampled outputs as cross-entropy targets to reshape the token-level decoding state, suppress long-tail trajectory drift in chain-of-thought generation, and consolidate both diversity and generalization.

\vspace{-0.5cm}

\subsection{Test-Time Adaptation}
Test-Time Adaptation (TTA) updates model parameters during inference to mitigate distributional shifts without costly offline retraining~\cite{liang2025comprehensive}, fundamentally blurring the boundary between training and deployment~\cite{park2025test}. Early TTA methods address the absence of ground-truth labels through self-supervised objectives, such as token-level entropy minimization~\cite{noori2026test} and next-token likelihood optimization~\cite{thrampoulidis2024implicit}, combined with selective parameter-efficient strategies including isolated layer updates, sparse weight adjustments, and low-rank parameter-efficient fine-tuning (PEFT) to scale gradient updates to billion-parameter models while preserving foundational linguistic capabilities~\cite{xu2026parameter}.

The optimization landscape of test-time fine-tuning increases risks of catastrophic forgetting and localized overfitting on single instances or small batches~\cite{luo2026dpc}. To mitigate these instabilities, recent work increasingly frames TTA through the lens of meta-learning and online optimization, leveraging principles from model-agnostic meta-learning~\cite{liang2025automatic} and Reptile~\cite{nichol2018first} to optimize initial pre-trained states for rapid downstream adaptation.

\vspace{-0.5cm}

\subsection{Training-Free Finetuning for Large Models}



Training-free tuning methods have garnered considerable attention as effective strategies for enhancing the inference-time performance of LLMs and VLMs without incurring additional computational overhead from parameter optimization. This kind of methods mainly explores token-level attention modulation for improving foundation model performance without any parameter modifications. These methods are based on the well-documented ``attention sink'' phenomenon, wherein a disproportionately large share of attention weight is systematically concentrated on the initial Beginning-of-Sequence (BOS) token~\citep{han2026zerotuning, xiao2024efficient, kaul2025attention, gu2025attention, barbero2025llms}. Attention sink motivates weight re-distribution techniques. Representative approaches such as Post-hoc Attention Steering (PASTA, \cite{zhang2024tell}) and AutoPASTA \cite{zhang2024model} identify and selectively amplify the attention weights assigned to task-relevant tokens, while Attention Calibration (ACT) suppresses non-initial sink tokens to redistribute attention more effectively~\cite{yu2024unveiling}. This principle has also been extended to vision-language models, where attention re-weighting toward visual tokens has been shown to effectively mitigate hallucinations~\cite{liu2024paying, zhu2025ibd}. More recently, Han et al. introduced ZeroTuning that applies head-specific attention adjustments to the BOS token, demonstrating consistent and generalizable performance improvements across foundation LLMs~\cite{han2026zerotuning}.

As Table \ref{tab:comparsion} shows, \texttt{Enlightenment} represents a new paradigm for training-free finetuning methods. Our opinion is that weights and architecture are the two cornerstones of a model. While existing training-free adaptations predominantly manipulate attention weights, \textbf{our work reveals that we can also re-allocate weight distribution from an architectural perspective. This architectural view opens up an orthogonal yet equally promising direction, which is to treat model distribution as a global structural property rather than a parametric one.} To the best of our knowledge, this view goes beyond weight-centric finetuning paradi-
gms, which was not explored before.

\vspace{-0.5cm}

\section{Methodology}

The overall framework of \texttt{Enlightenment} is illustrated in Figure~\ref{fig:overall}. The motivation behind our approach draws a loose conceptual analogy to neuroscience, where sudden insights in human learning are sometimes associated with rapid reorganization of neural pathways. We adopt this only as a high-level intuition, without claiming a direct biological basis. To reflect this notion in machine learning, we modify structures by either adjusting existing or establishing new shortcuts within pre-trained models, enabling them to self-improve their performance during inference without requiring additional post-training.

The unique characteristics of \texttt{Enlightenment} are that it requires no training data, no gradient computation, and no weight updates. It is built upon a unified principle: \textbf{modifying shortcuts for key modules/layers} by exerting scaling factors over them, while establishing new shortcuts can be thought as turning the scaling factor from 0 to 1 for otherwise absent paths. The standard residual connection offers no mechanism to control the \emph{mixing} between the residual stream and the sublayer output at inference time. Specifically, shortcuts are assigned with dynamic scaling parameters, ensuring the network can dynamically calibrate its attention to specific layers and more precisely govern the propagation strength of features. As a result, \texttt{Enlightenment} can enrich the representational capacity of large models without altering or tuning their pre-trained weights. What's more favorable is that this enhanced expressivity is generalizable across diverse tasks and model families. 

While the underlying principle is shared, the integration points and companion modules regarding shortcut modification could differ across different model families. From an information-flow perspective, multimodal networks require residual damping to suppress visual noise, while text-only models require attention modulation to redistribute linguistic focus~\cite{jiang2026latent,kim2016multimodal,xiao2024efficient}. Now, we elaborate on how we perform shortcut modification on VLMs and LLMs, respectively.


\vspace{-0.5cm}

\subsection{Enlightenment for LLM}
\vspace{-0.2cm}

LLMs frequently exhibit an ``attention sink'' phenomenon, wherein disproportional attention weight is systematically allocated to the initial BOS token~\citep{han2026zerotuning, xiao2024efficient, kaul2025attention, gu2025attention, barbero2025llms}. Beyond attention weights modification, our proposed method recalibrates BOS attention behavior by mixing heads with scaled shortcuts. This provides a per-head level recalibration rather than globally suppressing BOS attention. In this section, we first examine the basic mechanics of the attention in LLMs before detailing the \texttt{Enlightenment} integration.

\textbf{Attention mechanism}. Modern LLMs rely on the multi-head attention (MHA) mechanism to capture complex dependencies across tokenized sequences~\citep{vaswani2017attention}. We define an input sequence as a matrix $\mathbf{X}\in\mathbb{R}^{L\times d_{\text{model}}}$, where $L$ denotes the sequence length (total number of tokens) and $d_{\text{model}}$ represents the hidden state dimension.

To process information across distinct representation subspaces, the hidden dimension $d_{\text{model}}$ is split into $H$ parallel attention heads, each operating on a lower-dimensional feature space $d_h=d_{\text{model}}/H$. For a given attention head $h\in\{1,\dots,H\}$, the input activation sequence $\mathbf{X}$ is linearly projected into query ($\mathbf{Q}_h$), key ($\mathbf{K}_h$), and value ($\mathbf{V}_h$) matrices using learned projection weight matrix:
\begin{equation}
\mathbf{Q}_h=\mathbf{X}\mathbf{W}_q^h,\quad\mathbf{K}_h=\mathbf{X}\mathbf{W}_k^h,\quad\mathbf{V}_h=\mathbf{X}\mathbf{W}_v^h.
\end{equation}
Here, $\mathbf{W}_q^h,\mathbf{W}_k^h,\mathbf{W}_v^h\in\mathbb{R}^{d_{\text{model}}\times d_h}$ are head-specific projection matrices. The scaled dot-product attention for the $h$-th head maps the queries and keys to a valid probability distribution via the softmax operator, weighting the corresponding values accordingly:
\begin{equation}
\begin{aligned}\mathbf{A}_h&=\mathrm{softmax}\left(\frac{\mathbf{Q}_h\mathbf{K}_h^\top}{\sqrt{d_h}}\right)\in\mathbb{R}^{L\times L}\\\mathbf{O}_h&=\mathrm{AttnOutput}(\mathbf{Q}_h,\mathbf{K}_h,\mathbf{V}_h)=\mathbf{A}_h\mathbf{V}_h\in\mathbb{R}^{L\times d_h}.\end{aligned}
\label{eq:ah}
\end{equation}
In this formulation, $\mathbf{A}_h$ represents the post-softmax attention weight matrix, where each element $A_{h,q,k}$ defines the normalized alignment score between the $q$-th query token and the $k$-th key token. The factor $\sqrt{d_h}$ acts as a regularizing scaling variance to alleviate vanishing gradients during backpropagation.

The individual head outputs $\mathbf{O}_h$ are stacked across the head dimension to a tensor $\mathbf{O}\in\mathbb{R}^{H\times L\times d_h}$ (or batched as $\mathbb{R}^{B\times H\times L\times d_h}$). In a vanilla transformer, these heads are aggregated via a learned linear combination using the output projection weights $\mathbf{W}_o^h\in\mathbb{R}^{d_h\times d_{\text{model}}}$:
\begin{equation}
    \mathrm{MHA}(\mathbf{X})=\sum_{h=0}^H\mathbf{O}_h\mathbf{W}_o^h.
    \label{eq:mha}
\end{equation}

\begin{figure}[h]
    \centering
    \includegraphics{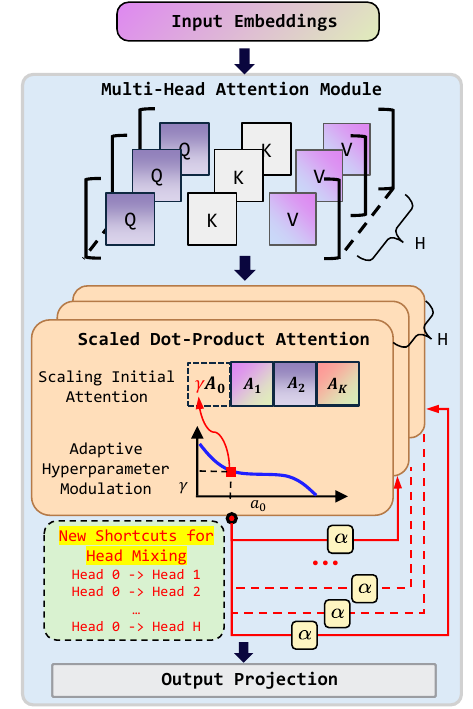}
    \caption{{The architectural framework of Enlightenment for LLMs.}}
    \vspace{-0.5cm}
    \label{fig:llm}
\end{figure}

\texttt{Enlightenment} modifies the attention head $\mathbf{A}_h$ across a designated subset of target layers $\mathcal{L}$ and heads $\mathcal{H}$ through head-to-head information mixing. By coupling the recalibration with the scattering shortcuts, the model successfully redistributes its linguistic focus. The overall framework is depicted in Figure~\ref{fig:llm}.

\textbf{Attention head mixing by inserting new shortcuts}. First, let $A_{b,h,q,k}$ denote the attention weight that the $q$-th query position assigns to the $k$-th key position in the $h$-th head. For a target layer $\ell \in \mathcal{L}$ and head $h \in \mathcal{H}$, leveraging the attention sink, we assign a scaling parameter $\gamma_{\ell,h} > 0$ to the initial attention matrix:
\begin{equation}
\begin{cases}
& \bar{A}_{b,h,q,0} = \gamma_{h} \cdot A_{b,h,q,0} \\
& \bar{A}_{b,h,q,\geq1} = A_{b,h,q,\geq1},
\end{cases}
\label{eq:bos_scaling}
\end{equation}
and normalize the values of newly scaled attention matrices $\bar{A}$ across all positions:
\begin{equation}
\tilde{{A}}_{b,h,q,:} = \frac{{\bar{A}}_{b,h,q,:}}{\sum_{k}\bar{A}_{b,h,q,k}}.
\label{eq:attnscale}
\end{equation}
According to Eq.~\eqref{eq:ah}, the output of the $h$-th attention head is
\begin{equation}
        {\mathbf{O}}_h = \tilde{\mathbf{A}}_h\mathbf{V}_h= [\tilde{\mathbf{A}}_{h,0}\mathbf{V}_{h,0}, \tilde{\mathbf{A}}_{h,1}\mathbf{V}_{h,1},\cdots,\tilde{\mathbf{A}}_{h,n}\mathbf{V}_{h,n}].
\end{equation}

After that, we establish new shortcuts that link the output of head 0 to all other target heads. For each edge from source head 0 to a target destination head $h$, we set a mixing rate $\alpha$ to control the additive injection. The matrix combining all heads' output is
\begin{equation}
       \tilde{\mathbf{O}} = [\tilde{\mathbf{A}}_0 \mathbf{V}_0, \alpha \tilde{\mathbf{A}}_0 \mathbf{V}_0 + \tilde{\mathbf{A}}_1 \mathbf{V}_1,
       \cdots , \alpha  \tilde{\mathbf{A}}_0 \mathbf{V}_0 +  \tilde{\mathbf{A}}_H \mathbf{V}_H].
\end{equation}
When $\alpha = 1$, the attention module remains unmodified. This process enriches the representation of the $h$-th head with contextual contributions from the initial head, which is a feature reuse. 

Finally, based on Eq.~\eqref{eq:mha}, the output of the MHA is
\begin{equation}
\begin{aligned}
        \mathrm{MHA}(\mathbf{X}) &=  \tilde{\mathbf{A}}_0 \mathbf{V}_0\mathbf{W}_{0}+ (\alpha \tilde{\mathbf{A}}_0 \mathbf{V}_0 + \tilde{\mathbf{A}}_1 \mathbf{V}_1)\mathbf{W}_{1}  + \\ &\cdots
    + (\alpha  \tilde{\mathbf{A}}_0 \mathbf{V}_0 +  \tilde{\mathbf{A}}_H \mathbf{V}_H)\mathbf{W}_H.
\end{aligned}
\label{eq:emha}
\end{equation}
These new shortcuts make the LLM dynamically redistribute linguistic focus from the initial attention block to other heads without introducing additional weights or training overhead.

\textbf{Adaptive hyperparameter modulation}. Hyperparameters $\Gamma=[\gamma_1, \gamma_2, \cdots, \gamma_H]$ and $\alpha$ are introduced in \texttt{Enlight-}
\texttt{enment}. Among them, $\alpha$ is a global parameter whose value is easy to determine by grid search. In contrast, in previous works, to alleviate the attention sink, initial heads exhibiting high attention values should be attenuated, while those with low attention should be amplified~\citep{han2026zerotuning, xiao2024efficient, kaul2025attention, gu2025attention, barbero2025llms}. Therefore, $\Gamma$ should be carefully calibrated for each head across layers to conform to this phenomenon. To this end, we introduce an adaptive strategy: 

$\small \bullet$ First, we calculate the value of the initial attention matrix of each head:
\begin{equation}
a_0^{(h)} = \frac{1}{B\cdot Q} \sum_{b=1}^B \sum_{q=1}^Q A_{b,h,q,0},
\label{eq:a0h}
\end{equation}
which serves as a runtime indicator of how strongly the $h$-th head attends to the initial position. We normalize $a_0^{(h)}$ to $[-1, 1]$ via
\begin{equation}
\hat{a}_0^{(h)} = 2 \cdot \frac{a_0^{(h)} - a_{\min}}{a_{\max} - a_{\min}} - 1, 
\label{eq:x0h}
\end{equation}
where $a_{\max}$ and $a_{\min}$ are the extreme values in $a^{(h)}$.

$\small \bullet$ Second, we aim at constructing a function $\gamma=f(\hat{a}_0)$. The general principle is that this function should be smooth over mid-range values, large for small attention values, and small for large attention values, thereby producing a more balanced scaling for attention values. To achieve this, we define four anchor pairs to ensure a basic shape for $f$. Each pair maps a high attention value to a lower $\gamma$, while a low initial attention value is amplified:
\begin{equation}
(\hat{a_0}, \gamma) \in \{(0.05, 1.5), (0.15, 1.1), (0.30, 1), (0.50, 0.7)\}.
\label{eq:pair}
\end{equation}
We fit these four anchor points to the Chebyshev polynomial function:
\begin{equation}
\gamma_h = T_d(x_h; \mathbf{c}) = \sum_{j=0}^{d} c_j \, T_j(x_h),
\end{equation}
with $\mathbf{c} = [c_0, \dots, c_d]$ are coefficients, and $T_j$ represents the $j$-th Chebyshev polynomial~\citep{islam2026bio}  satisfying
\begin{equation}
    T_0(x)=1, T_1(x)=x, T_{j}(x)=2x T_{j-1}(x) - T_{j-2}(x).
    \label{eq:chevb}
\end{equation}

\begin{figure}[h]
    \centering
\includegraphics{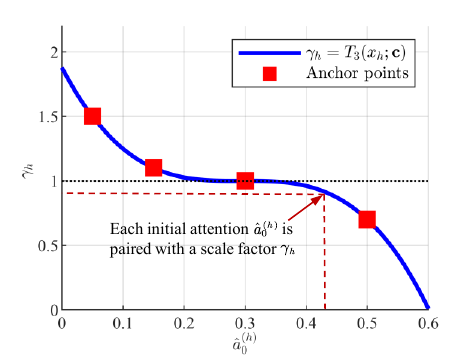}
    \caption{The fitted 3-order Chebyshev polynomial curve.}
\vspace{-0.5cm}
    \label{fig:curve}
\end{figure}

Figure \ref{fig:curve} presents a 3rd-order Chebyshev polynomial function.  $\gamma_h$ of each head can be determined by this function. 

Furthermore, we set $\beta$ as a task-dependent factor, which scales the Chebyshev polynomial $T_d$ horizontally.
\begin{equation}
    \gamma_h = \beta\cdot T_d(x_h; \mathbf{c}).
\end{equation}
Adjusting $\beta$ can effectively handle domain/task shift.

Table~\ref{tab:hyperparams} presents all the hyperparameters and their default values utilized in \texttt{Enlightenment} for LLMs.
\vspace{-0.5cm}

\begin{table}[ht]
\centering
\caption{Default hyper-parameters of the Enlightenment.}
\label{tab:hyperparams}
\begin{tabular}{@{}lcc@{}}
\toprule
Hyperparameters & Symbol & Value \\ \midrule
Polynomial degree    & $d$       & $3$ \\
Task-dependent factor          & $\beta$ & $1$ \\
Mixing rate      & $\alpha$  & $0.1$ \\
\bottomrule
\end{tabular}
\vspace{-1cm}
\end{table}

\subsection{Enlightenment for VLM}
\vspace{-0.2cm}

The core idea is to apply a single shared scalar to every residual connection within the Transformer decoder layers. Pre-trained sublayers, such as attention and Feed-Forward Networks, frequently become overly active during inference
\cite{you2026spark,korthikanti2023reducing}. This excessive activity leads to excessive information, introducing visual noise that degrades downstream predictive performance. By uniformly scaling down these sublayer outputs, the method effectively applies residual damping. This mechanism acts as a regularizer, suppressing superfluous sublayer activity while securely preserving the critical residual pathway that carries the core embedding information.

\begin{figure}[ht]
    \centering
    \includegraphics[width=\linewidth]{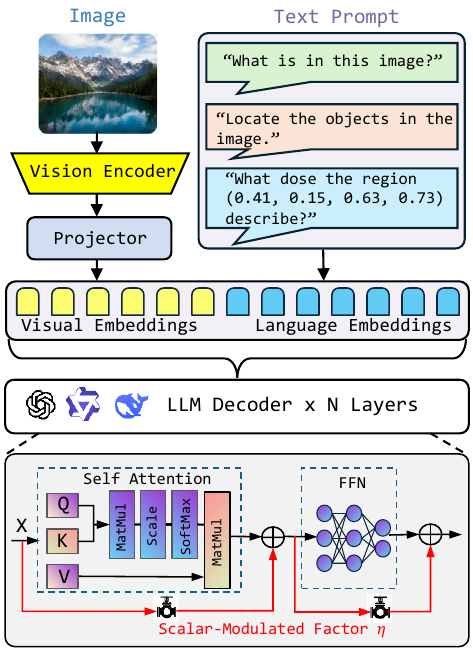}
    \caption{The architectural framework of Enlightenment for VLMs.}
    \vspace{-1cm}
    \label{fig:vlm}
\end{figure}

This method integrates scalar-modulated residual into the linear operations across layers, where a scaling hyperparameter can be adjusted to control contributions of these connections. The overall framework is illustrated in Figure~\ref{fig:vlm}.

\textbf{Scalar-modulated residuals}. The standard transformer decoder layer computes $x_{\ell+1} = x_\ell + F_\ell(\text{LN}(x_\ell))$, where $F_\ell$ is either self-attention or an FFN. We replace this with a single scalar-modulated variant:\vspace{-0.5cm}

\begin{equation}
\label{eq:epiphany}
x_{\ell+1} = x_\ell + \eta \cdot F_\ell(\text{LN}(x_\ell)),
\end{equation}
\vspace{-0.5cm}

where $\eta \in (0, 1]$ is a scalar shared across all $L$ decoder layers and both sublayer types (attention and FFN). At $\eta = 1$, the model operates in its unmodified state. At $\eta = 0.5$, the sublayer outputs are uniformly halved, introducing a form of \emph{residual damping} that mitigates noise while preserving the primary residual pathway carrying the original embedding information.
\vspace{-0.5cm}
\begin{table}[ht]
\centering
\caption{Structural comparison of VLM and LLM variants by Enlightenment.}
\label{tab:HN_structure}
\begin{tabular}{@{}p{2.5cm}
>{\columncolor[HTML]{DAE8FC}}p{2cm} 
>{\columncolor[HTML]{FFFFE4}}p{2cm} @{}}
\toprule
\textbf{Component}       & \cellcolor[HTML]{FFFFFF}\textbf{VLMs} & \cellcolor[HTML]{FFFFFF}\textbf{LLMs} \\ \midrule
Application point        & Decoder (Attn + FFN)                  & Per-head attention outputs            \\ \midrule
Model-specific companion & scalar-modulated residual shortcuts   & Head mixing shortcuts                 \\ \bottomrule
\end{tabular}
\vspace{-0.5cm}
\end{table}

The computational overhead is minimal, requiring only one scalar hyperparameter and one additional multiplication operation per layer. Consequently, \texttt{Enlightenment} requires no training data, no gradient computation, no weight updates. Our ablation studies demonstrate that $\eta$ can be easily tuned within the interval $[0.1, 1]$. Regarding the target layers, applying \texttt{Enlightenment} across all layers serves as a robust default strategy that consistently yields performance improvements.

In summary, Table~\ref{tab:HN_structure} compares the structural modifications applied to VLMs and LLMs by the proposed \texttt{Enlight-}
\texttt{enment} framework.

\vspace{-0.5cm}

\section{Experiments on LLMs}
\vspace{-0.2cm}

\subsection{Experimental Setup}
\vspace{-0.2cm}

\textbf{Models.} We evaluate \texttt{Enlightenment} on three models: Qwen3.5-9B-Base \citep{qwen3.5}, DeepSeek-R1-Distill-Qwen-14B \citep{deepseekai2025deepseekr1incentivizingreasoningcapability}, and Mimo-7B-RL~\cite{coreteam2025mimounlockingreasoningpotential}. Specifically, Qwen3.5 uses hybrid attention with 8 full-attention layers. DeepSeek-R1-14B is based on Qwen2 with 40 layers. Mimo-7B is a Qwen2-based model with 36 layers, GQA (32 query heads, 8 KV heads), 4096 hidden size, and SiLU activation, trained with RL. All results are test on 3 seeds (11, 22, 33) and compute the average values, with all decoder layers applied scalar-modulated factor for \texttt{Enlightenment}.

\begin{table*}[ht]
\caption{Accuracy (\%) of LLMs comparison. Best results per row in bold.}
\centering
\label{tab:llmresults}
\begin{tabular}{@{}lccccccccc@{}}
\toprule
                                   & \multicolumn{3}{c}{\textbf{Qwen3.5-9B}}                                                             & \multicolumn{3}{c}{\textbf{DeepSeek-R1-14B}}                                                                 & \multicolumn{3}{c}{\textbf{Mimo-7B-RL}}                                                             \\ \cmidrule(l){2-10} 
\multirow{-2}{*}{\textbf{Dataset}} & \textbf{Base}                & \textbf{ZT}                  & \textbf{Enlightenment}                & \textbf{Base}                & \textbf{ZT}                           & \textbf{Enlightenment}                & \textbf{Base}                & \textbf{ZT}                  & \textbf{Enlightenment}                \\ \midrule
SST-2                              & \cellcolor[HTML]{DAE8FC}92.2 & \cellcolor[HTML]{DAE8FC}92.3 & \cellcolor[HTML]{FFFFE4}\textbf{93.4} & \cellcolor[HTML]{DAE8FC}53.4 & \cellcolor[HTML]{DAE8FC}54.3          & \cellcolor[HTML]{FFFFE4}\textbf{59.2} & \cellcolor[HTML]{DAE8FC}90.7 & \cellcolor[HTML]{DAE8FC}90.6 & \cellcolor[HTML]{FFFFE4}\textbf{91.7} \\
CB                                 & \cellcolor[HTML]{DAE8FC}91.1 & \cellcolor[HTML]{DAE8FC}92.9 & \cellcolor[HTML]{FFFFE4}\textbf{93.9} & \cellcolor[HTML]{DAE8FC}79.2 & \cellcolor[HTML]{DAE8FC}\textbf{80.4} & \cellcolor[HTML]{FFFFE4}78.6          & \cellcolor[HTML]{DAE8FC}80.4 & \cellcolor[HTML]{DAE8FC}80.4 & \cellcolor[HTML]{FFFFE4}\textbf{82.4} \\
BoolQ                              & \cellcolor[HTML]{DAE8FC}88.2 & \cellcolor[HTML]{DAE8FC}89.3 & \cellcolor[HTML]{FFFFE4}\textbf{90.1} & \cellcolor[HTML]{DAE8FC}58.3 & \cellcolor[HTML]{DAE8FC}57.4          & \cellcolor[HTML]{FFFFE4}\textbf{65.8} & \cellcolor[HTML]{DAE8FC}71.5 & \cellcolor[HTML]{DAE8FC}70.6 & \cellcolor[HTML]{FFFFE4}\textbf{72.1} \\
ARCE                               & \cellcolor[HTML]{DAE8FC}94.0 & \cellcolor[HTML]{DAE8FC}94.1 & \cellcolor[HTML]{FFFFE4}\textbf{94.3} & \cellcolor[HTML]{DAE8FC}92.6 & \cellcolor[HTML]{DAE8FC}\textbf{92.7} & \cellcolor[HTML]{FFFFE4}\textbf{92.7} & \cellcolor[HTML]{DAE8FC}89.5 & \cellcolor[HTML]{DAE8FC}89.6 & \cellcolor[HTML]{FFFFE4}\textbf{90.8} \\
ARCC                               & \cellcolor[HTML]{DAE8FC}93.0 & \cellcolor[HTML]{DAE8FC}93.0 & \cellcolor[HTML]{FFFFE4}\textbf{94.3} & \cellcolor[HTML]{DAE8FC}87.3 & \cellcolor[HTML]{DAE8FC}\textbf{87.3} & \cellcolor[HTML]{FFFFE4}87.0          & \cellcolor[HTML]{DAE8FC}84.6 & \cellcolor[HTML]{DAE8FC}84.6 & \cellcolor[HTML]{FFFFE4}\textbf{86.6} \\
CQA                                & \cellcolor[HTML]{DAE8FC}54.7 & \cellcolor[HTML]{DAE8FC}54.7 & \cellcolor[HTML]{FFFFE4}\textbf{55.5} & \cellcolor[HTML]{DAE8FC}75.9 & \cellcolor[HTML]{DAE8FC}\textbf{77.3} & \cellcolor[HTML]{FFFFE4}76.7          & \cellcolor[HTML]{DAE8FC}76.6 & \cellcolor[HTML]{DAE8FC}76.6 & \cellcolor[HTML]{FFFFE4}\textbf{77.8} \\
PIQA                               & \cellcolor[HTML]{DAE8FC}49.3 & \cellcolor[HTML]{DAE8FC}50.6 & \cellcolor[HTML]{FFFFE4}\textbf{51.4} & \cellcolor[HTML]{DAE8FC}85.1 & \cellcolor[HTML]{DAE8FC}\textbf{86.3} & \cellcolor[HTML]{FFFFE4}85.0          & \cellcolor[HTML]{DAE8FC}82.6 & \cellcolor[HTML]{DAE8FC}83.5 & \cellcolor[HTML]{FFFFE4}\textbf{83.9} \\ \midrule
Avg.                               & \cellcolor[HTML]{DAE8FC}80.4 & \cellcolor[HTML]{DAE8FC}81.0 & \cellcolor[HTML]{FFFFE4}\textbf{81.8} & \cellcolor[HTML]{DAE8FC}76.0 & \cellcolor[HTML]{DAE8FC}76.5          & \cellcolor[HTML]{FFFFE4}\textbf{77.9} & \cellcolor[HTML]{DAE8FC}82.3 & \cellcolor[HTML]{DAE8FC}82.3 & \cellcolor[HTML]{FFFFE4}\textbf{83.6} \\
Avg. $\Delta$ vs base              & \cellcolor[HTML]{DAE8FC}---  & \cellcolor[HTML]{DAE8FC}+0.6 & \cellcolor[HTML]{FFFFE4}+1.5          & \cellcolor[HTML]{DAE8FC}---  & \cellcolor[HTML]{DAE8FC}+0.6          & \cellcolor[HTML]{FFFFE4}+1.9          & \cellcolor[HTML]{DAE8FC}---  & \cellcolor[HTML]{DAE8FC}+0.0 & \cellcolor[HTML]{FFFFE4}+1.3          \\ \bottomrule
\end{tabular}
\vspace{-0.5cm}
\end{table*}

In addition, model compression is increasingly useful in the era of LLMs. In the field of model compression, the performance degradation needs to be controlled carefully. Therefore, we also evaluate if \texttt{Enlightenment} works for the compressed LLM. Now, in industry, the mainstream model compression is based on quantization. We evaluate the \textbf{W4A4C2} quantization on the DeepSeek-R1-Distill-Qwen-1.5B checkpoint. Specifically, the base model is a full-precision FP16 baseline, while the \textbf{W4A4C2} quantization model uses W4 weight dequantization (4-bit restored), A4 activation fake-quantization via LobCQ (4-bit, group size 128, outlier ratio 0.01), and C2 KV-cache vector quantization (2-bit K/V, group size 32, residual length 32).

\textbf{Datasets.} Our evaluation spans several standard benchmarks. For classification, we include SST-2 (Stanford Sentiment Treebank; sentiment analysis; 2 classes) \citep{socher-etal-2013-recursive}, CB (CommitmentBank; natural language inference; 3 classes) \citep{demarneffe2019commitmentbank}, and BoolQ (Boolean Questions; yes/no QA; 2 classes) \citep{clark-etal-2019-boolq}. For multiple-choice and multi-class tasks, we use ARCE (AI2 Reasoning Challenge Easy; multiple-choice science QA; 4 classes) \citep{clark2018think}, ARCC (AI2 Reasoning Challenge Challenge; multiple-choice science QA; 4 classes) \citep{clark2018think}, CQA (CommonsenseQA; multiple-choice commonsense QA; 5 classes) \citep{talmor-etal-2019-commonsenseqa}, and PIQA (Physical Interaction QA; multiple-choice physical commonsense QA; 2 classes) \citep{bisk2020piqa}. Across all datasets, we adopt perplexity-based scoring over label templates. For compressed models, all results use the vendored lm-evaluation-harness~\footnote{\url{https://github.com/EleutherAI/lm-evaluation-harness}}. Tasks are evaluated in a zero-shot, log-likelihood-based multiple-choice setting with batch size 8, max length 2048, and no bootstrapping.

\textbf{Baselines.} We compare \texttt{Enlightenment} against ZeroTuning (ZT)~\citep{han2026zerotuning}. This method is the most recent SOTA training-free tuning method for LLMs, serving as an updated version of earlier methods~\cite{zhang2024tell,zhang2024model,yu2024unveiling}. We therefore adopt it as the sole baseline for comparison. For \texttt{Enlightenment}, we apply attention head mixing with uniform per-head rates. All methods include no training and are evaluated at inference time. Specifically, for the W4A4C2 quantization model, we apply the \texttt{Enlightenment} attention head mixing to all layers and all heads at a base rate of $\alpha=0.1, \beta=0.5$.

\vspace{-0.5cm}

\subsection{Main Results}
\vspace{-0.2cm}

Table \ref{tab:llmresults} compares the accuracy of three LLMs across seven benchmarks. Overall, \texttt{Enlightenment} achieves the best average performance for all three models, improving Qwen3.5-9B from 80.4\% to 81.8\%, DeepSeek-R1-14B from 76.0\% to 77.9\%, and Mimo-7B-RL from 82.3\% to 83.6\%. Compared with the base setting, \texttt{Enlightenment} yields average gains of +1.5, +1.9, and +1.3 points, respectively, while ZT provides smaller or negligible improvements. At the dataset level, Enlightenment obtains the highest accuracy in most cases, particularly showing clear gains on SST-2, BoolQ, and ARCC. Although ZT slightly outperforms Enlightenment on some tasks for DeepSeek-R1-14B, such as CB, CQA, and PIQA, overall, \texttt{Enlightenment} provides a more consistent and stronger improvement across models and datasets.
\vspace{-0.8cm}
\begin{table}[ht]
\centering
\caption{Using Enlightenment (all layers/heads) after quantization LLM.}
\label{tab:zt}
\begin{tabular}{@{}lc|cc@{}}
\toprule
\textbf{Tasks} & \textbf{Base FP16}            & \textbf{W4A4C2}               & \textbf{+Enlightenment}                \\ \midrule
BoolQ          & \cellcolor[HTML]{EFEFEF}0.686 & \cellcolor[HTML]{DAE8FC}0.378 & \cellcolor[HTML]{FFFFE4}\textbf{0.567} \\
RTE            & \cellcolor[HTML]{EFEFEF}0.603 & \cellcolor[HTML]{DAE8FC}0.527 & \cellcolor[HTML]{FFFFE4}\textbf{0.534} \\
WinoGrande     & \cellcolor[HTML]{EFEFEF}0.556 & \cellcolor[HTML]{DAE8FC}0.496 & \cellcolor[HTML]{FFFFE4}\textbf{0.509} \\
ARCE           & \cellcolor[HTML]{EFEFEF}0.617 & \cellcolor[HTML]{DAE8FC}0.251 & \cellcolor[HTML]{FFFFE4}\textbf{0.261} \\
PIQA           & \cellcolor[HTML]{EFEFEF}0.652 & \cellcolor[HTML]{DAE8FC}0.495 & \cellcolor[HTML]{FFFFE4}\textbf{0.528} \\
MMLU           & \cellcolor[HTML]{EFEFEF}0.366 & \cellcolor[HTML]{DAE8FC}0.229 & \cellcolor[HTML]{FFFFE4}\textbf{0.244} \\ \midrule
Average        & \cellcolor[HTML]{EFEFEF}0.580 & \cellcolor[HTML]{DAE8FC}0.396 & \cellcolor[HTML]{FFFFE4}\textbf{0.441} \\ \bottomrule
\end{tabular}
\vspace{-0.5cm}
\end{table}

\begin{table*}[ht]
\centering
\caption{Comparison of perplexity-based marginal improvement. ``Margin'' = $\log P(\text{correct}) - \log P(\text{wrong})$, higher is better.}
\label{tab:margin_examples}
\begin{tabular}{@{}p{7cm}cccc|c@{}}
\toprule
\textbf{Sentence (partial)}                                     & \multicolumn{2}{c}{\textbf{Baseline}}                             & \multicolumn{2}{c}{\textbf{Enlightenment}}                       &                                     \\ \cmidrule(lr){2-5}
                                                                & $\log P(\text{corr})$           & $\log P(\text{wrong})$          & $\log P(\text{corr})$           & $\log P(\text{wrong})$          & \multirow{-2}{*}{\textbf{$\Delta$}} \\ \midrule
\texttt{the subtle strength of ``elling'' is that it never...} (pos) & \cellcolor[HTML]{DAE8FC}$-2.59$ & \cellcolor[HTML]{DAE8FC}$-6.03$ & \cellcolor[HTML]{FFFFE4}$-2.59$ & \cellcolor[HTML]{FFFFE4}$-6.34$ & \cellcolor[HTML]{FFCCC9}$+0.31$     \\
\texttt{this riveting WWII moral suspense story deserves...} (neg)   & \cellcolor[HTML]{DAE8FC}$-2.80$ & \cellcolor[HTML]{DAE8FC}$-4.56$ & \cellcolor[HTML]{FFFFE4}$-2.78$ & \cellcolor[HTML]{FFFFE4}$-4.78$ & \cellcolor[HTML]{FFCCC9}$+0.23$     \\
\texttt{the socio-histo-political treatise is told in earn...} (pos) & \cellcolor[HTML]{DAE8FC}$-2.30$ & \cellcolor[HTML]{DAE8FC}$-5.47$ & \cellcolor[HTML]{FFFFE4}$-2.28$ & \cellcolor[HTML]{FFFFE4}$-5.66$ & \cellcolor[HTML]{FFCCC9}$+0.20$     \\
\texttt{cool ? ...} (pos)                                            & \cellcolor[HTML]{DAE8FC}$-2.23$ & \cellcolor[HTML]{DAE8FC}$-4.25$ & \cellcolor[HTML]{FFFFE4}$-2.14$ & \cellcolor[HTML]{FFFFE4}$-4.34$ & \cellcolor[HTML]{FFCCC9}$+0.19$     \\ \bottomrule
\end{tabular}
\end{table*}

\begin{table*}[ht]
\caption{Full output of a CommonsenseQA example.}
\centering
\label{tab:full_output1}
\begin{tabular}{@{}lp{8cm}cc@{}}
\toprule
\textbf{Question}                        & \texttt{After he got hired he hoped for success . What does he need to do?} & \multicolumn{2}{c}{\textbf{Log-probability}}                                                   \\ \midrule
\textbf{True answer}                     & \multicolumn{1}{l|}{B (have ambition)}                                                   & \textbf{Baseline}                          & \textbf{Enlightenment}                            \\ \midrule
                                         & \multicolumn{1}{l|}{\cellcolor[HTML]{EFEFEF}A (work)}                                    & \cellcolor[HTML]{DAE8FC}$\log P = {-5.97}$ & \cellcolor[HTML]{FFFFE4}$\log P = -6.06$          \\
                                         & \multicolumn{1}{l|}{\cellcolor[HTML]{EFEFEF}\textbf{B (have ambition)}}                  & \cellcolor[HTML]{DAE8FC}$\log P = -5.97$   & \cellcolor[HTML]{FFFFE4}\textbf{$\log P = -6.00$} \\
                                         & \multicolumn{1}{l|}{\cellcolor[HTML]{EFEFEF}C (talk)}                                    & \cellcolor[HTML]{DAE8FC}$\log P = -11.50$  & \cellcolor[HTML]{FFFFE4}$\log P = -11.56$         \\
                                         & \multicolumn{1}{l|}{\cellcolor[HTML]{EFEFEF}D (go shopping)}                             & \cellcolor[HTML]{DAE8FC}$\log P = -10.56$  & \cellcolor[HTML]{FFFFE4}$\log P = -10.69$         \\
\multirow{-5}{*}{\textbf{Choice scores}} & \multicolumn{1}{l|}{\cellcolor[HTML]{EFEFEF}E (eat)}                                     & \cellcolor[HTML]{DAE8FC}$\log P = -10.25$  & \cellcolor[HTML]{FFFFE4}$\log P = -10.38$         \\ \midrule
\textbf{Baseline prediction}             & \multicolumn{3}{c}{\cellcolor[HTML]{DAE8FC}A (work) — incorrect}                                                                                                                          \\
\textbf{Enlightenment prediction}        & \multicolumn{3}{c}{\cellcolor[HTML]{FFFFE4}B (have ambition) — correct}                                                                                                                   \\ \bottomrule
\end{tabular}
\vspace{-0.5cm}
\end{table*}

Table~\ref{tab:zt} demonstrates that \texttt{Enlightenment} consistently improves the performance of the quantized LLM across all evaluated tasks. Under the aggressive W4A4C2 quantization setting, the average score drops substantially from 0.580 in FP16 to 0.396, indicating a clear degradation caused by low-bit quantization. Applying \texttt{Enlightenment} after quantization raises the average score to 0.441, yielding a +0.045 absolute improvement over W4A4C2. The gain is especially pronounced on BoolQ, where performance increases from 0.378 to 0.567, recovering a large portion of the accuracy lost during quantization. Improvements are also observed on RTE, WinoGrande, ARCE, PIQA, and MMLU, showing that the benefit is not limited to a single benchmark. These results suggest that \texttt{Enlightenment} can effectively mitigate quantization-induced degradation and enhance the robustness of low-bit LLM inference without restoring the model to full precision.

We also provide qualitative examples comparing baseline and \texttt{Enlightenment} on the SST-2 sentiment classification task using Qwen3.5-9B-Base. The model is prompted with 

\texttt{Sentence: ...} 

\texttt{Answer: ...}

While individual greedy-decoded outputs are often identical, the improvement from Enlightenment operates on the log-probability margin between the correct and incorrect label. Table~\ref{tab:margin_examples} shows four examples from the SST-2 validation set where Enlightenment increases the log-probability margin for the correct label. The ``Margin'' column shows the difference $\log P(\text{correct}) - \log P(\text{wrong})$; larger is better.

In each case, the Enlightenment suppresses the log-probab-
ility of the wrong label (more negative) while marginally increasing or maintaining the correct label probability, widening the margin by $+0.19$ to $+0.31$ log-probability units. 

Moreover, Table~\ref{tab:full_output1} shows a CommonsenseQA example. CommonsenseQA is a 5-way multiple-choice task. The evaluation computes $\log P(\text{answer letter})$ for each of the five choices and selects the highest. Under the baseline, the model assigns nearly identical log-probabilities to choices A and B (both $-5.97$), and the tie is resolved in favor of the wrong answer. Enlightenment shifts the distribution: choice B's score rises relative to choice A ($A$ from $-5.97$ to $-6.06$, $B$ from $-5.97$ to $-6.00$), producing a clear margin of $0.06$\,nats in favor of the correct answer. The remaining choices (C, D, E) are unaffected because the BOS scaling targets only the top-ranking candidates.

\vspace{-0.5cm}

\subsection{Hyperparameter Sensitivity}

\vspace{-0.2cm}

We evaluate the sensitivity of hyperparameters in the \texttt{Enligh-}
\texttt{tenment} by varying each hyperparameter in isolation while keeping all others at their defaults. The evaluation is based on the BoolQ task (100 samples, seed 11). Table~\ref{tab:sensitivity} reports accuracy across two hyperparameters, polynomial degree ($d$) and mixing strength ($\alpha$). As the task-dependent factor $\beta$ can be used to finetune the model in different tasks, we set it to 1 in this experiment.

\vspace{-0.5cm}

\begin{table}[ht]
\centering
\caption{Hyperparameter sensitivity on BoolQ (Qwen3.5-9B, 100 $\times$ 1). Default values are bold-faced.}
\label{tab:sensitivity}
\begin{tabular}{@{}llc@{}}
\toprule
\textbf{Section}                                                & \textbf{Value} & \textbf{Acc. (\%)} \\ \midrule
\rowcolor[HTML]{DAE8FC} 
\cellcolor[HTML]{DAE8FC}                                        & \textbf{3}     & \textbf{94.0}      \\
\rowcolor[HTML]{DAE8FC} 
\cellcolor[HTML]{DAE8FC}                                        & 1              & 92.0               \\
\rowcolor[HTML]{DAE8FC} 
\cellcolor[HTML]{DAE8FC}                                        & 2              & 93.0               \\
\rowcolor[HTML]{DAE8FC} 
\multirow{-4}{*}{\cellcolor[HTML]{DAE8FC}Ploynomial degree $d$} & 5              & 94.0               \\ \midrule
\rowcolor[HTML]{FFFFE4} 
\cellcolor[HTML]{FFFFE4}                                        & \textbf{0.1}   & \textbf{94.0}      \\
\rowcolor[HTML]{FFFFE4} 
\cellcolor[HTML]{FFFFE4}                                        & 1 (disabled)   & 91.0               \\
\rowcolor[HTML]{FFFFE4} 
\cellcolor[HTML]{FFFFE4}                                        & 0.05           & 94.0               \\
\rowcolor[HTML]{FFFFE4} 
\cellcolor[HTML]{FFFFE4}                                        & 0.2            & 94.0               \\
\rowcolor[HTML]{FFFFE4} 
\multirow{-5}{*}{\cellcolor[HTML]{FFFFE4}Mixing $\alpha$}       & 0.5            & 94.0               \\ \bottomrule
\end{tabular}
\vspace{-0.5cm}
\end{table}

The hyperparameter sensitivity results on BoolQ show that the method is generally robust to changes in both the polynomial degree $d$ and the mixing coefficient $\alpha$. Using the default setting $d=3$ achieves the best accuracy of 94.0\%, while lower degrees lead to slight drops, with $d=1$ reaching 92.0\% and $d=2$ reaching 93.0\%. This suggests that a moderate polynomial degree is sufficient to capture the desired effect, and increasing it further does not provide additional gains. For the mixing coefficient, the default value $\alpha=0.1$ also achieves 94.0\% accuracy, and performance remains unchanged across different settings. In contrast, disabling mixing by setting $alpha=1$ reduces accuracy to 91.0\%, indicating that the mixing mechanism is important for maintaining strong performance.
\vspace{-0.5cm}

\subsection{Ablation Study}

\textbf{Adaptive hyperparameter modulation.} We use Qwen3.5-9B to conduct the ablation study. Table~\ref{tab:component_ablation} evaluates the individual and combined impact of our proposed mechanisms by isolating the head mixing component (HeadMix only) and applying adaptive hyperparameter modulation (Adap.+
HeadMix). First, applying HeadMix yields highly inconsistent results across tasks, leading to a minor average regression to 80.3\% compared to the Base model's 80.4\%. This instability indicates that while head mixing can capture specialized attention patterns, it introduces optimization variability. In contrast, the joint configuration (Adap.+HeadMix) successfully mitigates these vulnerabilities, securing the highest performance on six out of seven datasets and achieving a superior overall average accuracy of 81.8\%. The adaptive hyperparameter modulation framework acts as a vital stabilizer, reversing the isolated regressions of HeadMix on SST-2 and CQA to produce net gains over Base (+1.2\% and +0.8\%, respectively). This strong synergy is particularly pronounced on reasoning-heavy datasets like CB (+2.8\%) and PIQA (+2.1\%).

\vspace{-0.5cm}

\begin{table}[ht]
\caption{Accuracy (\%) of component ablation on Qwen3.5-9B.}
\centering
\label{tab:component_ablation}
\begin{tabular}{@{}lccc@{}}
\toprule
\textbf{Datasets} & \textbf{Baseline}                         & \textbf{HeadMix only}                 & \textbf{Adap.+HeadMix}                         \\ \midrule
SST-2   & \cellcolor[HTML]{EFEFEF}92.2 & \cellcolor[HTML]{DAE8FC}90.0 & \cellcolor[HTML]{FFFFE4}93.4          \\
CB      & \cellcolor[HTML]{EFEFEF}91.1 & \cellcolor[HTML]{DAE8FC}92.9 & \cellcolor[HTML]{FFFFE4}93.9          \\
BoolQ   & \cellcolor[HTML]{EFEFEF}88.2 & \cellcolor[HTML]{DAE8FC}91.0 & \cellcolor[HTML]{FFFFE4}90.1          \\
ARCE    & \cellcolor[HTML]{EFEFEF}94.0 & \cellcolor[HTML]{DAE8FC}93.0 & \cellcolor[HTML]{FFFFE4}94.3          \\
ARCC    & \cellcolor[HTML]{EFEFEF}93.0 & \cellcolor[HTML]{DAE8FC}93.0 & \cellcolor[HTML]{FFFFE4}94.3          \\
CQA     & \cellcolor[HTML]{EFEFEF}54.7 & \cellcolor[HTML]{DAE8FC}53.0 & \cellcolor[HTML]{FFFFE4}55.5          \\
PIQA    & \cellcolor[HTML]{EFEFEF}49.3 & \cellcolor[HTML]{DAE8FC}49.5 & \cellcolor[HTML]{FFFFE4}51.4          \\ \midrule
Avg.    & \cellcolor[HTML]{EFEFEF}80.4 & \cellcolor[HTML]{DAE8FC}80.3 & \cellcolor[HTML]{FFFFE4}\textbf{81.8} \\ \bottomrule
\end{tabular}
\vspace{-0.5cm}
\end{table}

\textbf{Different layer and head ranges.} We conduct experiments to analyze the impact of applying \texttt{Enlightenment} in different layer and head ranges. The results are presented in Table~\ref{tab:layers}. For target layers \(\mathcal{L}\), it indicates that our method is highly robust to the choice of target layers, since all tested layer configurations, ranging from full-layer intervention to sparse subsets, consistently achieve \(94.0\%\) accuracy. This suggests that a small number of selected layers is already sufficient to capture the relevant effect. For target heads \(\mathcal{H}\), the trend is similarly stable: most partial head selections yield \(94.0\%\), while using all heads provides a slightly higher accuracy of \(95.0\%\). Overall, these findings show that the approach is not sensitive to precise layer/head selection.
\vspace{-0.5cm}

\begin{table}[ht]
\centering
\caption{Changing Enlightenment layers and heads on BoolQ (Qwen3.5-9B, 100 $\times$ 1). }
\label{tab:layers}
\begin{tabular}{@{}llc@{}}
\toprule
\textbf{Section}                                                      & \textbf{Value}                          & \textbf{Acc. (\%)} \\ \midrule
\rowcolor[HTML]{DAE8FC} 
\cellcolor[HTML]{DAE8FC}                                              & Full layers                             & 94.0               \\
\rowcolor[HTML]{DAE8FC} 
\cellcolor[HTML]{DAE8FC}                                              & \{3,7,15,23,31\}                        & 94.0               \\
\rowcolor[HTML]{DAE8FC} 
\cellcolor[HTML]{DAE8FC}                                              & \{3,7,11,15,19,23,27,31\}               & 94.0               \\
\rowcolor[HTML]{DAE8FC} 
\cellcolor[HTML]{DAE8FC}                                              & \{7,15,23\}                             & 94.0               \\
\rowcolor[HTML]{DAE8FC} 
\multirow{-5}{*}{\cellcolor[HTML]{DAE8FC}Target layers $\mathcal{L}$} & \{23,27,31\}                            & 94.0               \\
\rowcolor[HTML]{DAE8FC} 
                                                                      & \{15,23\}                               & 94.0               \\ \midrule
\rowcolor[HTML]{FFFFE4} 
\cellcolor[HTML]{FFFFE4}                                              & \{0,\dots,7\} (full) & \textbf{95.0}      \\
\rowcolor[HTML]{FFFFE4} 
\cellcolor[HTML]{FFFFE4}                                              & \{0,8\}                                 & 94.0               \\
\rowcolor[HTML]{FFFFE4} 
\cellcolor[HTML]{FFFFE4}                                              & \{0\}                                   & 94.0               \\
\rowcolor[HTML]{FFFFE4} 
\cellcolor[HTML]{FFFFE4}                                              & \{8\}                                   & 94.0               \\
\rowcolor[HTML]{FFFFE4} 
\multirow{-5}{*}{\cellcolor[HTML]{FFFFE4}Target heads $\mathcal{H}$}  & \{0,4,8,12\}                            & 94.0               \\ \bottomrule
\end{tabular}
\vspace{-1cm}
\end{table}

\begin{table*}[ht]
\centering
\caption{Accuracy (\%) of compared post-tuning methods across 5 benchmarks. Best results per row in bold.}
\label{tab:method_compvlm}
\begin{tabular}{@{}lcccccccccc@{}}
\toprule
\textbf{}          & \textbf{}            & \multicolumn{3}{c}{\textbf{MiMo-VL-7B}}                                                                & \multicolumn{3}{c}{\textbf{Qwen2-VL-7B}}                                                                                 & \multicolumn{3}{c}{\textbf{Qwen3-VL-8B}}                                                                        \\ \midrule
\textbf{Dataset}   & \textbf{\# Samples}  & \textbf{Base}                 & \textbf{ZT}                   & \textbf{Enlightenment}                 & \textbf{Base}                          & \textbf{ZT}                            & \textbf{Enlightenment}                 & \textbf{Base}                 & \textbf{ZT}                            & \textbf{Enlightenment}                 \\ \midrule
MMBench            & 300                  & \cellcolor[HTML]{DAE8FC}84.33 & \cellcolor[HTML]{DAE8FC}84.35 & \cellcolor[HTML]{FFFFE4}\textbf{85.67} & \cellcolor[HTML]{DAE8FC}79.33          & \cellcolor[HTML]{DAE8FC}79.67          & \cellcolor[HTML]{FFFFE4}\textbf{85.33} & \cellcolor[HTML]{DAE8FC}92.00 & \cellcolor[HTML]{DAE8FC}91.67          & \cellcolor[HTML]{FFFFE4}\textbf{92.00} \\
MME                & 1,187                & \cellcolor[HTML]{DAE8FC}43.22 & \cellcolor[HTML]{DAE8FC}43.23 & \cellcolor[HTML]{FFFFE4}\textbf{51.22} & \cellcolor[HTML]{DAE8FC}88.21          & \cellcolor[HTML]{DAE8FC}88.29          & \cellcolor[HTML]{FFFFE4}\textbf{89.13} & \cellcolor[HTML]{DAE8FC}84.08 & \cellcolor[HTML]{DAE8FC}84.33          & \cellcolor[HTML]{FFFFE4}\textbf{85.15} \\
AI2D               & 3,088                & \cellcolor[HTML]{DAE8FC}78.59 & \cellcolor[HTML]{DAE8FC}78.61 & \cellcolor[HTML]{FFFFE4}\textbf{79.05} & \cellcolor[HTML]{DAE8FC}50.87          & \cellcolor[HTML]{DAE8FC}50.55          & \cellcolor[HTML]{FFFFE4}\textbf{53.04} & \cellcolor[HTML]{DAE8FC}33.55 & \cellcolor[HTML]{DAE8FC}33.55          & \cellcolor[HTML]{FFFFE4}\textbf{45.53} \\
RealworldQA        & 765                  & \cellcolor[HTML]{DAE8FC}72.55 & \cellcolor[HTML]{DAE8FC}72.51 & \cellcolor[HTML]{FFFFE4}\textbf{73.20} & \cellcolor[HTML]{DAE8FC}\textbf{66.41} & \cellcolor[HTML]{DAE8FC}\textbf{66.41} & \cellcolor[HTML]{FFFFE4}66.27          & \cellcolor[HTML]{DAE8FC}69.02 & \cellcolor[HTML]{DAE8FC}\textbf{70.20} & \cellcolor[HTML]{FFFFE4}69.80          \\
MMMU               & 541                  & \cellcolor[HTML]{DAE8FC}26.80 & \cellcolor[HTML]{DAE8FC}26.81 & \cellcolor[HTML]{FFFFE4}\textbf{27.54} & \cellcolor[HTML]{DAE8FC}26.25          & \cellcolor[HTML]{DAE8FC}26.62          & \cellcolor[HTML]{FFFFE4}\textbf{28.10} & \cellcolor[HTML]{DAE8FC}34.01 & \cellcolor[HTML]{DAE8FC}34.20          & \cellcolor[HTML]{FFFFE4}\textbf{35.49} \\ \midrule
\multicolumn{2}{l}{Avg.}                  & \cellcolor[HTML]{DAE8FC}61.10 & \cellcolor[HTML]{DAE8FC}61.10 & \cellcolor[HTML]{FFFFE4}\textbf{63.34} & \cellcolor[HTML]{DAE8FC}62.21          & \cellcolor[HTML]{DAE8FC}62.31          & \cellcolor[HTML]{FFFFE4}\textbf{64.37} & \cellcolor[HTML]{DAE8FC}62.53 & \cellcolor[HTML]{DAE8FC}62.79          & \cellcolor[HTML]{FFFFE4}\textbf{65.59} \\
\multicolumn{2}{l}{Avg. $\Delta$ vs base} & \cellcolor[HTML]{DAE8FC}---   & \cellcolor[HTML]{DAE8FC}+0.00 & \cellcolor[HTML]{FFFFE4}\textbf{+2.24} & \cellcolor[HTML]{DAE8FC}---            & \cellcolor[HTML]{DAE8FC}+0.09          & \cellcolor[HTML]{FFFFE4}\textbf{+2.16} & \cellcolor[HTML]{DAE8FC}---   & \cellcolor[HTML]{DAE8FC}+0.26          & \cellcolor[HTML]{FFFFE4}\textbf{+3.06} \\ \bottomrule
\end{tabular}
\vspace{-0.5cm}
\end{table*}

\section{Experiments on VLMs}

\subsection{Experimental Setup}

\textbf{Models}: We evaluate 3 VLMs: MiMo-VL-7B~\citep{coreteam2025mimounlockingreasoningpotential}, Qwen2-VL-7B~\citep{wang2024qwen2vl}, Qwen3-VL-8B~\citep{bai2025qwen3} that were widely used in recent two years. All models are loaded at BF16 precision with eager attention. The experiments are conducted on NVIDIA RTX PRO 6000 GPUs (96\,GiB VRAM each). For each evaluation task, a fresh copy of the model is loaded, and the compared method is applied before running the benchmark.

\textbf{Benchmarks}: We use 5 benchmarks that cover the mainstream multimodal tasks. MMBench~\citep{liu2024mmbench} (4,876 natural-image MCQ), MME~\citep{fu2026mme} (2,374 Yes/No perception), AI2D~\citep{kembhavi2016ai2d} (3,088 diagram MCQ), RealworldQA~\citep{realworldqa2024} (765 spatial reasoning MCQ), MMMU (541 Architecture~\&~Engineering MCQ via HuggingFace datasets)~\citep{yue2024mmmu}.

\textbf{Methods}. Similar to LLM experiments, we compare our method with the ZeroTuning (ZT)~\citep{han2026zerotuning} that uses the original uniform BOS scaling. For \texttt{Enlightenment}, we apply a single shared scalar modulator $\gamma=0.5$ to all residual connections ($x + \gamma \cdot F$), patching both attention and MLP sublayers on every decoder layer. 

\textbf{Inference details}. All benchmarks use greedy decoding. For standard models, we set {\tt Max New Tokens}=10 for MCQ (MMBench, AI2D, MMMU) and Yes/No (MME), $20$ for free-form RealworldQA. For MiMo-VL-7B---a reasoning model that generates chain-of-thought via {\tt <think>} tags, we set {\tt Max New Tokens}=256 for MCQ and $64$ for free-form to accommodate deliberation. Answer extraction for MCQ checks the first option letter appearing in the response for Qwen models or the last option letter after {\tt </think>} for the MiMo model. MME matches response substrings against ``yes''/``no'' ground truth.

\vspace{-0.5cm}

\subsection{Main Results}

Table~\ref{tab:method_compvlm} compares the accuracy of three settings, Base, ZT, and \texttt{Enlightenment}, across five multimodal benchmarks on MiMo-VL-7B, Qwen2-VL-7B, and Qwen3-VL-8B. Overall, \texttt{Enlightenment} consistently achieves the strongest performance, yielding the highest average accuracy across all three models. Compared with the Base, \texttt{Enlightenment} improves average accuracy by +2.24\% for MiMo-VL-7B, +1.92\% for Qwen2-VL-7B, and +3.06\% for Qwen3-VL-8B. These gains are particularly pronounced on challenging benchmarks, such as MME for MiMo-VL-7B, where accuracy increases over 8\%, and AI2D for Qwen3-VL-8B, where performance rises substantially from 33.55\% to 45.53\%. In contrast, ZT provides only marginal improvement. Although \texttt{Enlightenment} is not the top performer in every individual case, it achieves the best or tied-best results on most datasets and consistently improves the overall average. These results demonstrate that \texttt{Enlightenment} is a more effective and robust post-tuning strategy than ZT, delivering clear and stable gains across different VLM architectures and benchmark types.

Furthermore, we conduct per-category analysis. The primary evaluation results demonstrate that \texttt{Enlightenment} achieves competitive performance on the MMBench dataset when integrated with the Qwen2-VL-7B model. Since MMBench is a comprehensive benchmark, we conduct a per-category analysis to characterize the specific strengths and limitations of this method. Table~\ref{tab:mmbench_cat} breaks down the MMBench performance gains by L3 category (spanning 20 categories across the full 1,292-sample set). Notably, \texttt{Enlight}-\texttt{enment} yields the most substantial improvements on perception and recognition tasks: image emotion ($+16.67\%$), attribute recognition ($+6.67\%$), and image style ($+6.45\%$). At the L2 aggregation level, ``coarse perception'' improves by $+6.08\%$, while ``fine-grained perception (instance-level)'' increases by $+4.50\%$. Conversely, performance declines in relational reasoning categories: nature\_relation ($-11.29\%$), future\_prediction ($-4.00\%$), and the L2 aggregate ``relation reasoning'' drops by $1.15\%$. This category-level performance profile explains different benchmark-level generalization abilities of \texttt{Enlightenment}: it benefits MMBench because the dataset is dominated by perception tasks, whereas showing a slight degrades performance on RealworldQA due to its inherent focus on relational reasoning.
\vspace{-0.5cm}

\begin{table}[ht]
\centering
\caption{{Per-category MMBench accuracy deltas} with Enlightenment on Qwen2-VL-7B (1292 samples). Top 5 gainers and losers.}
\label{tab:mmbench_cat}
\scalebox{0.9}{
\begin{tabular}{@{}lccc@{}}
\toprule
\textbf{Category}      & \textbf{Baseline}               & \textbf{Enlightenment}          & $\boldsymbol{\Delta}$                     \\ \midrule
\multicolumn{4}{c}{\textit{Top gainers (L2: coarse perception $+6.08$, finegrained instance $+4.50$)}}                                 \\
image\_emotion         & \cellcolor[HTML]{DAE8FC}68.33\% & \cellcolor[HTML]{FFFFE4}85.00\% & \cellcolor[HTML]{FFCCC9}\textbf{$+16.67$} \\
attribute\_recognition & \cellcolor[HTML]{DAE8FC}88.33\% & \cellcolor[HTML]{FFFFE4}95.00\% & \cellcolor[HTML]{FFCCC9}\textbf{$+6.67$}  \\
image\_style           & \cellcolor[HTML]{DAE8FC}87.10\% & \cellcolor[HTML]{FFFFE4}93.55\% & \cellcolor[HTML]{FFCCC9}\textbf{$+6.45$}  \\
social\_relation       & \cellcolor[HTML]{DAE8FC}85.48\% & \cellcolor[HTML]{FFFFE4}91.94\% & \cellcolor[HTML]{FFCCC9}\textbf{$+6.45$}  \\
celebrity\_recognition & \cellcolor[HTML]{DAE8FC}92.93\% & \cellcolor[HTML]{FFFFE4}98.99\% & \cellcolor[HTML]{FFCCC9}\textbf{$+6.06$}  \\ \midrule
\multicolumn{4}{c}{\textit{Top losers (L2: relation reasoning $-1.15$)}}                                                               \\
function\_reasoning    & \cellcolor[HTML]{DAE8FC}91.67\% & \cellcolor[HTML]{FFFFE4}90.00\% & \cellcolor[HTML]{FFCCC9}$-1.67$           \\
attribute\_comparison  & \cellcolor[HTML]{DAE8FC}78.43\% & \cellcolor[HTML]{FFFFE4}76.47\% & \cellcolor[HTML]{FFCCC9}$-1.96$           \\
future\_prediction     & \cellcolor[HTML]{DAE8FC}80.00\% & \cellcolor[HTML]{FFFFE4}76.00\% & \cellcolor[HTML]{FFCCC9}$-4.00$           \\
nature\_relation       & \cellcolor[HTML]{DAE8FC}88.71\% & \cellcolor[HTML]{FFFFE4}77.42\% & \cellcolor[HTML]{FFCCC9}\textbf{$-11.29$} \\ \bottomrule
\end{tabular}}
\vspace{-0.5cm}
\end{table}

We also provide qualitative examples comparing Baseline and \texttt{Enlightenment} on the MMBench dataset. The model is Qwen2-VL-7B-Instruct (Qwen2VL architecture, 28 decoder layers, hidden dim 3584, 8.29B parameters), and loaded at BF16 precision, greedy decoding, 10 generated tokens.

\begin{figure}[ht]
\centering
\includegraphics{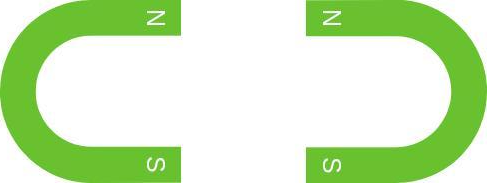}
\caption{Two magnets placed with the same poles facing each other.}
\label{fig:example}
\vspace{-0.5cm}
\end{figure}

\texttt{Question:} Two magnets are placed as shown in the Figure~\ref{fig:example} (both sets of arrows point in the same direction inside each magnet, indicating the same pole orientation). Will these magnets attract or repel each other?

\texttt{Options:}
\begin{itemize}
    \item \texttt{A. Repel.}
    \item \texttt{B. Attract.}
\end{itemize}

\texttt{Correct answer: A (Repel.)}

The model output is presented in Table~\ref{tab:example_output}. The baseline model incorrectly predicts ``Attract'' despite the poles being aligned in the same direction. \texttt{Enlightenment}'s residual damping suppresses the over-active sublayer outputs that contribute to this visual reasoning error, allowing the model to correctly identify the physical relationship. This example is representative of the pattern we observe across the MMBench dataset: \texttt{Enlightenment} improves perceptual and physical reasoning accuracy while preserving the model's overall knowledge.

\vspace{-0.5cm}

\begin{table}[ht]
\centering
\caption{Baseline vs. Enlightenment output.}
\label{tab:example_output}
\begin{tabular}{@{}lp{5cm}@{}}
\toprule
\textbf{Method} & \textbf{Model output} \\
\midrule
Baseline & {\tt B. Attract.} $\times$ \\
Enlightenment ($\gamma=0.5$) & {\tt A. Repel.} $\checkmark$ \\
\bottomrule
\end{tabular}
\vspace{-0.5cm}
\end{table}

\vspace{-0.5cm}

\subsection{Sensitivity Analysis}
\label{sec:gamma_sweep}

\begin{figure}[ht]
    \centering
    \includegraphics{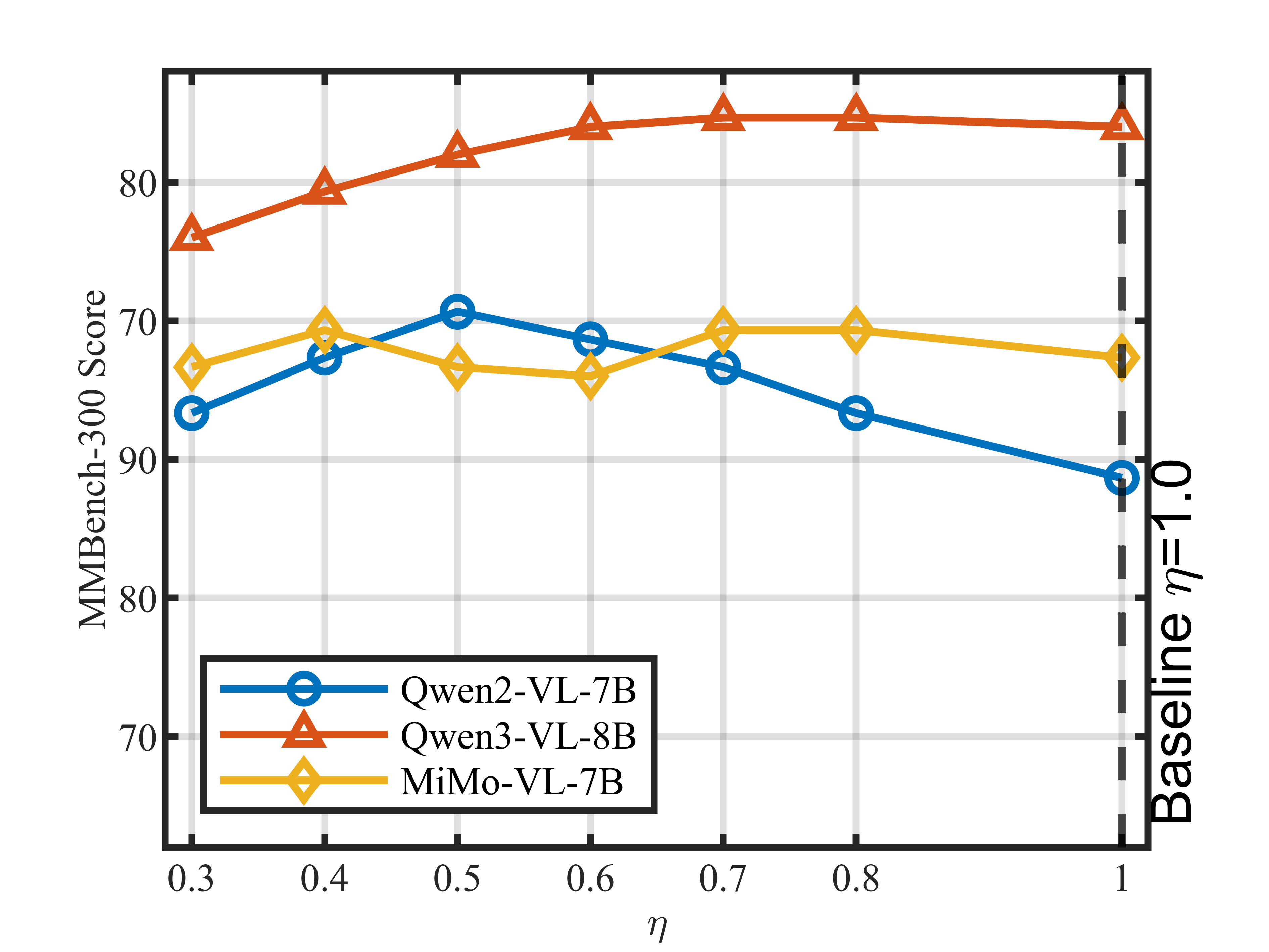}
    \caption{Sweep $\eta$ on MMBench-300 across 4 primary models. $\eta=1.0$ is the baseline.}
    \vspace{-0.5cm}
    \label{fig:gamma_sweep}
\end{figure}

\textbf{Hyperparameters}. \texttt{Enlightenment} introduces only a single hyperparameter $\eta$. To evaluate its impact, we conduct a sensitivity analysis of $\eta$ on MMBench-300, as summarized in Figure~\ref{fig:gamma_sweep}. Most models exhibit stable performance across different values of $\eta$. Qwen2-VL-7B is the sole exception, displaying a clear inverted-U pattern that peaks at $\eta=0.5$ with a 6\% performance gain. In contrast, the performance of Qwen2-VL-2B, Qwen3-VL-8B, and MiMo-VL-7B are subjected to slight changes across the interval $\eta \in [0.4, 0.8]$. This highlights that \texttt{Enlightenment} is highly robust to hyperparameter selection, making it an ideal plug-and-play, zero-tuning approach to boost the performance of base VLMs.


\textbf{Different layer ranges}. We conduct experiments to analyze the impact of applying \texttt{Enlightenment} in different layer ranges. These results are summarized in Table~\ref{tab:layer_range}. First, the MMBench results show that gating any single layer range in isolation degrades performance; early layers are particularly sensitive (-12\%), while late layers also experience a decline (-7\%). Conversely, results on the AI2D dataset reveal a {different pattern}: gating only the middle layers (L10--19) improves accuracy by 4.9\%, suggesting that visual diagrammatic reasoning benefits from mid-layer residual damping even in isolation. On the other hand, late-layer gating on AI2D causes the most severe performance drop among all configurations, indicating that diagram comprehension relies heavily on late-layer representations. Uniformly gating all layers yields an improvement of 1.49\% on AI2D, consistent with the full-benchmark evaluation. Overall, an all-layer application is necessary to achieve general improvements across different benchmarks, and targeted layer selection offers a promising avenue for task-specific fine-tuning to achieve optimal performance.
\vspace{-0.5cm}

\begin{table}[ht]
\centering
\caption{Apply Enlightenment to different layer-range on Qwen2-VL-7B with $\eta=0.5$ . $\Delta$ from baseline.}
\label{tab:layer_range}
\scalebox{0.9}{
\begin{tabular}{@{}lcccc@{}}
\toprule
\textbf{Modulated Layers} & \textbf{MMBench}                         & \textbf{MMBench $\boldsymbol{\Delta}$}   & \textbf{AI2D}                            & \textbf{AI2D $\boldsymbol{\Delta}$}      \\ \midrule
None (baseline)           & \cellcolor[HTML]{DAE8FC}81.67\%          & \cellcolor[HTML]{DAE8FC}---              & \cellcolor[HTML]{FFFFE4}51.55\%          & \cellcolor[HTML]{FFFFE4}---              \\
Early (L0--9)             & \cellcolor[HTML]{DAE8FC}69.00\%          & \cellcolor[HTML]{DAE8FC}$-12.67$         & \cellcolor[HTML]{FFFFE4}52.65\%          & \cellcolor[HTML]{FFFFE4}$+1.10$          \\
Middle (L10--19)          & \cellcolor[HTML]{DAE8FC}76.33\%          & \cellcolor[HTML]{DAE8FC}$-5.34$          & \cellcolor[HTML]{FFFFE4}\textbf{56.48\%} & \cellcolor[HTML]{FFFFE4}\textbf{$+4.92$} \\
Late (L20--27)            & \cellcolor[HTML]{DAE8FC}74.33\%          & \cellcolor[HTML]{DAE8FC}$-7.34$          & \cellcolor[HTML]{FFFFE4}46.44\%          & \cellcolor[HTML]{FFFFE4}$-5.12$          \\
All (L0--27)              & \cellcolor[HTML]{DAE8FC}\textbf{84.00\%} & \cellcolor[HTML]{DAE8FC}\textbf{$+2.33$} & \cellcolor[HTML]{FFFFE4}53.04\%          & \cellcolor[HTML]{FFFFE4}$+1.49$          \\ \bottomrule
\end{tabular}}
\vspace{-1cm}
\end{table}

\vspace{-0.2cm}

\subsection{Ablation Study}
\vspace{-0.2cm}

Our method uses \texttt{Enlightenment} on both attention and FFN layers. We conduct experiments to show their individual contributions. 

\vspace{-0.5cm}

\begin{table}[ht]
\centering
\caption{Attention vs. FFN gating on Qwen2-VL-7B. Best per benchmark in \textbf{bold}. $\Delta$ is from baseline.}
\label{tab:attn_ffn}
\begin{tabular}{@{}lcccc@{}}
\toprule
\textbf{Config}      & ${\eta_{\text{attn}}}$ & ${\eta_{\text{ffn}}}$ & \textbf{MMBench} & ${\Delta}$       \\ \midrule
\rowcolor[HTML]{EFEFEF} 
Baseline             & 1.0                    & 1.0                   & 81.67\%          & ---              \\
\rowcolor[HTML]{DAE8FC} 
Attn only            & 0.5                    & 1.0                   & 63.67\%          & $-18.00$         \\
\rowcolor[HTML]{DAE8FC} 
FFN only             & 1.0                    & 0.5                   & 56.33\%          & $-25.33$         \\
\rowcolor[HTML]{FFFFE4} 
Both (Enlightenment) & 0.5                    & 0.5                   & \textbf{84.00\%} & \textbf{$+2.33$} \\ \bottomrule
\end{tabular}
\vspace{-0.5cm}
\end{table}
Table~\ref{tab:attn_ffn} reveals that {neither attention-only nor FFN-only gating works in isolation}. Scaling only attention degrades MMBench by approximately -18\%, and scaling only FFN degrades MMBench by about -25\%. Only when both sublayer types are suppressed together does the accuracy improves on MMBench. This demonstrates that applying \texttt{Enlightenment} to attention and FFN residuals forms a coupled system: the scalar modulation's effectiveness derives from uniform suppression of both sublayer types simultaneously.

\vspace{-0.5cm}

\section{Discussion}

\vspace{-0.2cm}

\subsection{Swapping Enlightenment Strategies Between VLM and LLM}
\vspace{-0.2cm}

In this section, we examine why distinct \texttt{Enlightenment} strategies are needed for VLMs and LLMs. We evaluate both HeadMix and scalar-modulated shortcuts on two baseline models: Qwen2-VL-7B (a VLM) and Mimo-7B-RL (an LLM). The results are reported in Table~\ref{tab:llmvsvlm1}.

For the VLM, HeadMix degrades performance, lowering MMBench accuracy from 79.3\% to 76.5\%. In contrast, the scalar-modulated residual achieves 85.3\%, a gain of 6.0 percentage points over the baseline. For the LLM, the pattern reverses: scalar-modulated residuals cause accuracy to drop to 89.5\%, indicating that this mechanism, despite its effectiveness in multimodal contexts, does not benefit text-only reasoning. HeadMix raises accuracy to 91.7\%, confirming its alignment with the structural characteristics of LLMs.
\vspace{-0.5cm}

\begin{table}[ht]
\centering
\caption{HeadMix vs. Scalar-Modulated Shortcuts on VLM and LLM.}
\label{tab:llmvsvlm1}
\scalebox{0.9}{
\begin{tabular}{@{}llcc@{}}
\toprule
\textbf{Models}                                                                                                 & \textbf{Method}                          & \textbf{Dataset} & \textbf{Acc. \%}                \\ \midrule
\rowcolor[HTML]{DAE8FC} 
\cellcolor[HTML]{DAE8FC}                                                                               & Baseline                        & MMBench & 79.3                   \\
\rowcolor[HTML]{DAE8FC} 
\cellcolor[HTML]{DAE8FC}                                                                               & HeadMix                         & MMBench & 76.5 ($-2.8$)          \\
\rowcolor[HTML]{DAE8FC} 
\multirow{-3}{*}{\cellcolor[HTML]{DAE8FC}\begin{tabular}[c]{@{}l@{}}Qwen2-VL-\\ 7B (VLM)\end{tabular}} & Scalar-Modulated ($\eta{=}0.5$) & MMBench & \textbf{85.3 ($+6.0$)} \\ \midrule
\rowcolor[HTML]{FFFFE4} 
\cellcolor[HTML]{FFFFE4}                                                                               & Baseline                        & SST-2 & 90.7                  \\
\rowcolor[HTML]{FFFFE4} 
\cellcolor[HTML]{FFFFE4}                                                                               & HeadMix                         & SST-2 & \textbf{91.7($+1.0$)}         \\
\rowcolor[HTML]{FFFFE4} 
\multirow{-3}{*}{\cellcolor[HTML]{FFFFE4}\begin{tabular}[c]{@{}l@{}}Mimo-\\ 7B-RL (LLM)\end{tabular}}  & Scalar-Modulated ($\eta{=}0.5$) & SST-2 & 89.5 ($-1.2$) \\ \bottomrule
\end{tabular}}
\vspace{-0.5cm}
\end{table}

Taken together, these findings reveal a clear cross-architec-ture asymmetry: the two \texttt{Enlightenment} strategies exhibit pronounced \textbf{model-type specificity}. HeadMix introduces new scaling and additive computation within the attention module, which proves beneficial for LLMs yet disrupts the inherently more complex multimodal attention patterns that jointly process text and visual tokens, thereby degrading VLM performance. Conversely, scalar-modulated residuals regulate feature flow through a lightweight gating mechanism that selectively attenuates extraneous signals entering the transformer decoder, reducing cross-modal noise and yielding stable, substantial gains for VLMs. This reciprocal behavior underscores the necessity of tailoring \texttt{Enlighten-}
\texttt{ment} strategies to each model family.

\vspace{-0.5cm}

\subsection{Output Distributions}
\vspace{-0.2cm}

We use the {Qwen3.5-9B-Base} model with the prompt \textit{``Classify the sentence into one of the following categories: positive or negative. Sentence: A breathtaking masterpiece of modern cinema. Answer: ...''}. A forward hook was registered on the output self-attention module of the 15$^{th}$ layer (one of the target attention layers) to capture the post-softmax attention weight tensor of the shape $[1, 16, 28, 28]$ (batch, head groups, query length, key length). 16 head groups correspond to the KV-head dimension after GQA expansion in Qwen3.5 (32 query heads, 8 KV heads, with the query heads grouped into 16 tracked slices by the patched attention function). The ``BOS attention'' for each head group is computed as $\frac{1}{Q} \sum_{q=0}^{Q-1} \A_{h,q,0}$, \textit{i.e.}, the mean attention weight that each query position assigns to the BOS key position ($k=0$), averaged over all $Q=28$ query positions.

Specifically, we note that generic scaling rate assigns $0.68, 0.68, 0.66, 0.86, 0.51$ to the 3$^{rd}$, 7$^{th}$, 15$^{th}$, 23$^{th}$, and 31$^{st}$ layers. All five rates are below $1.0$, meaning that suppressing the initial attention at target layers yields better accuracy on tasks. This is consistent with the ``attention sink'' phenomenon, where pretrained transformers allocate excessive attention to the initial token, and reducing this baseline improves task-specific focus. 

\vspace{-0.5cm}

\begin{figure}[ht]
\centering
\includegraphics{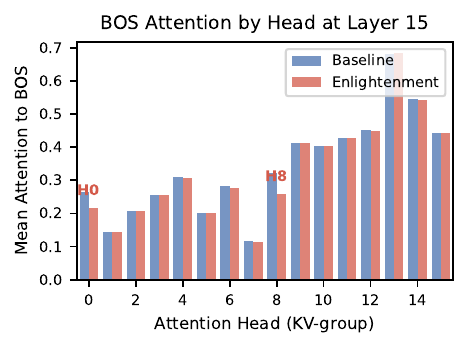}
\caption{The BOS attention by the head at the 15$^{th}$ layer. {Enlightenment} suppresses BOS attention only at heads 0 and 8.}
\vspace{-0.5cm}
\label{fig:bos_attention}
\end{figure}

Moreover, Figure~\ref{fig:bos_attention} illustrates the effect of \texttt{Enlighten-}
\texttt{ment} on BOS attention across heads in a targeted layer. Unlike previous methods that globally reduce initial attention values, \texttt{Enlightenment} selectively decreases BOS attention only on heads 0 and 8, while other heads remain slightly decreased or nearly unchanged. This phenomenon indicates that the injection of new shortcuts operates through a mechanism distinct from attention-based modification. Our method can dynamically control attention across heads rather than globally suppressing BOS attention.

\vspace{-0.7cm}

\subsection{Activation Analysis}
\vspace{-0.2cm}

\begin{figure}[ht]
    \centering
    \includegraphics[scale=1.1]{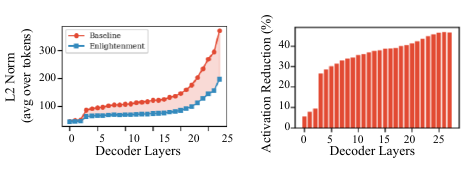}
    \caption{Comparison of the baseline and Enlightenment VLM decoder activation outputs. }
    \vspace{-0.5cm}
    \label{fig:saliency}
\end{figure}

To directly verify the over-active sublayer hypothesis in VLM, we compare the hidden state norms of all 28 decoder layers on the same input. First, Figure~\ref{fig:saliency} present the L2 Norm of decoder layers and actication reduction by \texttt{Enlightenment}. We have the following findings.  {(1) Baseline activations explode in late layers.} The L2 norm grows from 47 at layer 0 to 371 at layer 27, which is a nearly 8$\times$ increase. This suggests that sublayer outputs in deep layers accumulate redundant information rather than refining the residual stream. {(2) \texttt{Enlightenment} suppresses late layers proportionally more.} Layers 0--2 see only $5$--$10\%$ reduction, while layers 20--27 see $40$--$47\%$ reduction, consistent with the baseline trend that over-activation concentrates at depth. 
\vspace{-0.7cm}

\begin{figure}[ht]
    \centering
    \includegraphics{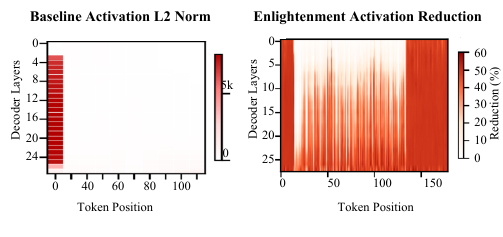}
    \caption{Comparison of the baseline and Enlightenment VLM decoder activation outputs correspondence to token position. Note that, for the baseline, we visualize only the preceding 100+ token positions because most attention values are concentrated at the initial token position.}
    \label{fig:saliency2}
\end{figure}

Moreover, we present the VLM decoder activation outputs correspondence to token position in Figure~\ref{fig:saliency2}. The results confirm the ``attention sink'' phenomenon in the baseline model, where the initial token dominates the attention values. In contrast, our scalar-modulated method successfully suppresses this effect, substantially reducing the initial token's outsized activation. Finally, the activation profile directly supports the mechanism: pretrained sublayer outputs become increasingly noisy at depth, and a single $\eta$ suppresses that excess while preserving the residual path.
\vspace{-0.5cm}

\vspace{-1cm}

\section{Conclusion}

In this paper, we have introduced \texttt{Enlightenment}, a novel, training-free finetuning paradigm designed to facilitate zero-shot self-improvement in large-scale models. Drawing inspiration from the neurological phenomenon of enlightenment during resting states, our method modifies shortcuts within pre-trained transformers without requiring any data, gradient computation, or weight update. \texttt{Enlightenment} effectively regularizes over-active sublayer outputs and optimizes token-level attention allocation. Extensive empirical evaluations across multiple VLMs and LLMs demonstrate that \texttt{Enlightenment} consistently elevates baseline performance across a wide array of multimodal, perception, and reasoning benchmarks. Moreover, the method exhibits a markedly low sensitivity for hyperparameters. Our work confirms the promising trajectory of ``NeuroAI'' principles. Future work will be digging deeper into the brain's mechanisms to boost the self-improvement of large models.








\vspace{-0.5cm}

\bibliographystyle{spbasic}
\bibliography{main}

\end{document}